# LostNet: A smart way for lost and find

Meihua Zhou[1], Ivan Fung[2], Li Yang[1], Nan Wan[1], Keke Di[1]

Tingting Wang*

[1] School of Medical Information, Wannan Medical Collage, Wuhu City, China

{mhzhou, 20120037, wannan, mhzhou0412}@wnmc.edu.cn

[2] work.ivanfung@gmail.com

* School of Medical Information, Wannan Medical Collage, Wuhu City, China

wangtt@wnmc.edu.cn

**Abstract:**

Due to the enormous population growth of cities in recent years, objects are frequently lost and unclaimed on public transportation, in restaurants, or any other public areas. While services like Find My iPhone can easily identify lost electronic devices, more valuable objects cannot be tracked in an intelligent manner, making it impossible for administrators to reclaim a large number of lost and found items in a timely manner. We present a method that significantly reduces the complexity of searching by comparing previous images of lost and recovered things provided by the owner with photos taken when registered lost and found items are received. In this research, we will primarily design a photo matching network by combining the fine-tuning method of MobileNetv2 with CBAM Attention and using the Internet framework to develop an online lost and found image identification system. Our implementation gets a testing accuracy of 96.8% using only 665.12M GLFOPs and 3.5M training parameters. It can recognize practice images and can be run on a regular laptop.

**Introduction:**

The population density and the quantity of lost objects are both rising in the areas where urban rail transit is located around the world at the present time; yet, the traditional manual search service is ineffective. In this situation, there is an immediate need to speed up the development of intelligent lost and found systems in order to lessen the difficulty that transportation operators face when it comes to lost and found management. Deep learning can be used to build recognition and classification models for lost and found items. This is a new method that can reduce reliance on manual labour, quickly and accurately identify categories, significantly reduce human service costs for transportation operators, and better practice green development strategies.

The application of convolutional neural networks is widespread throughout many scientific subfields pertaining to image recognition and classification (Sun et al., 2021). It is not a passing fad that academics in a wide variety of fields have shifted their attention to the study and practical use of image recognition. In the realm of trash classification, a method for garbage image classification was devised. It was built on an enhanced version of MobileNet v2 and paired with transfer learning to increase the real-time performance and accuracy of garbage image classification models (Huang et al., 2021).

**Research Background:**

Many academics are also engaged in questioning the status quo in the many subfields that make up the area of applied image recognition. (Yang et al., 2015) attempting to identify plant leaves by the utilization of a hierarchical model that is based on CNN. A study on establishing the optimal size of the training data set that is required to achieve high classification accuracy with low variance in medical image classification systems is presented by Cho et al. (2015). (Purnama et al., 2019) offer a method for the classification and diagnosis of skin diseases that is suitable for use in teledermatology.

It has also been demonstrated that transfer learning is beneficial in a variety of contexts. Convolutional neural networks are used in the methodology that Lee et al. (2016) propose as a fine-grained classification method for large-scale plankton databases. The implementation of transfer learning in CNN is one potential solution. (Liu et al., 2020) apply unsupervised transfer learning to CNN training to address these problems. Specifically, they transform similarity learning into deep ordinal classification with the assistance of several CNN experts who were pretrained over large-scale-labelled everyday image sets. These CNN experts jointly determine image similarities and provide pseudo labels for classification. (Purwar et al., 2020) make use of some models that are related to convolutional neural networks (CNNs) in order to identify mesangial hypercellularity in MEST-C. (Herzog et al., 2021) concentrate on the classification of MRI for the diagnosis of early and progressive dementia by utilizing transfer learning architectures that employ Convolutional Neural Networks-CNNs, as a base model, and fully connected layers of SoftMax functions or Support Vector Machines-SVMs. (Phankokkruad, 2021) propose the three CNN models for detecting lung cancer using VGG16, ResNet50V2, and DenseNet201 architectures. The proposed method is based on transfer learning.

**Methodology:**

In a procedure in which CNN is being used to address more and more components of the problem, the lost and found problem has not been solved in any intelligent method as of yet. As a result, we recommend taking a methodical approach by utilizing Mobilnet v2 and an intuitive graphical user interface (GUI). In this study, we build on earlier research to further investigate the detection and categorization of lost and discovered items, and we present an approach that combines perceptual hashing with MobileNet v2 transfer learning.

In order to solve the relatively complex problems associated with the image dataset of lost and found items, we carried out extensive research, classified the most common items that are lost and found into ten categories using questionnaires and market research, and produced private dataset images using crawlers, real-world photography, and research examples. After that, the generated dataset is used to train the network and establish an intelligent recognition classification model for lost and found images. This model solves the problems of labor cost and time cost consumption that are associated with conventional methods and proposes an accurate and complete solution with scientific and accurate experimental data.

I. Mobilenet v2

An outstanding example of a good lightweight convolutional neural network is MobileNetV2 (Sandler et al., 2018). The network creates a reverse residual and linear bottlenecks, both of which are helpful for feature extraction; the linear activation that is used in the final layer of the inverted residual structure prevents the loss of low-latitude information; and the traditional convolution is replaced by depth-separable convolution, which significantly reduces the amount of the model's calculations as well as the number of the parameters that it uses. It was developed specifically for pictures and has applications in image categorization as well as the development of generic features.

Convolution on a depth-wise and point-wise scale are the two components that make up depth-separable convolution. The depth separable convolution algorithm is not the same as the standard convolution algorithm. During the process of convolution, each channel of the feature map is covered by exactly one convolution nucleus, and the total number of convolution nuclei is equal to the total number of channels. The following phrases can be used to describe depth convolution:

$$O_{x,y,c} = \sum_{w,h}^{W,H} K_{w,h,c} \cdot I_{x+w,y+h,c} \quad (\text{formula1})$$

In the equation presented above, the variable O stands for the output feature graph, c stands for the channel of the feature graph, x and y stand for the coordinates of the output feature graph on channel, K stands for the convolutional kernel with wide W and high H, I stand for the input feature graph, and w and h stand for the convolutional kernel weight element coordinates of channel.

The primary difference between point-by-point and standard convolution is the size of the convolution kernel, which in point-by-point convolution is fixed at 1x1. The first step of depth separable convolution is to employ depth convolution to extract the characteristics of each channel. Next, point-by-point convolution is used to correlate the extracted channel characteristics. The depth separable convolution is intended to reduce the number of parameters and computations required by the conventional convolution. When compared to the total number of calculations involved in the traditional convolution, the results are as follows:

$$\frac{R_1}{R_2} = \frac{D_f^2 D_k^2 I + D_f^2 IO}{D_f^2 D_k^2 IO} = \frac{1}{N} + \frac{1}{D_k^2} \quad (\text{formula2})$$

In the formula 2 shown above: R1 and R2 represent the calculations of depth separable convolution and standard convolution, respectively; Df and Dk represent the height and width of the input feature matrix; I represent the depth of the input feature matrix; and O represents the depth of the output feature matrix.

In the process of feature extraction, MobileNetV2 makes use of a depth detachable convolution with a size of 3x3, which means that the calculation cost is 1/9 of what it would be for a standard convolution. However, the reduction in accuracy is very minimal, which is also one of the most notable qualities of the model.

Figure 1 presents the organizational structure of the MobileNetV2 network. It is primarily made up of three components: the front end is a convolutional neural network (CNN), which is constructed by several layers of convolution, and then the average pooling of 1,280 7x7 blocks is utilized to generate 1,280 nerves. Element, which was afterwards completely coupled with one thousand neurons.

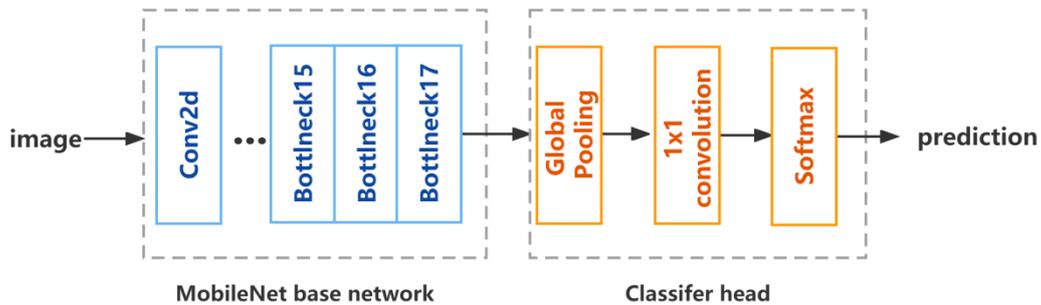

Figure1 MobileNet v2 based classifier

MobileNet v2 borrows ResNet's residual connection notion and uses it to suggest a reverse residual structure, as seen in the picture. The structure employs PW convolution to increase dimension, 3x3 DW convolution to extract channel characteristics, then PW convolution once again to decrease feature dimension. Notably, as seen in Figure 2, a residual link exists only when the step duration is 1. In the second situation, when the step size is 2, the series structure is employed directly, and the reverse residual structure is upgraded and then lowered, so that the network permits smaller input and output dimensions, hence decreasing the amount of computation and parameters of the network. Simultaneously, residual connection may increase the effectiveness of gradient propagation and deepen the network layer.

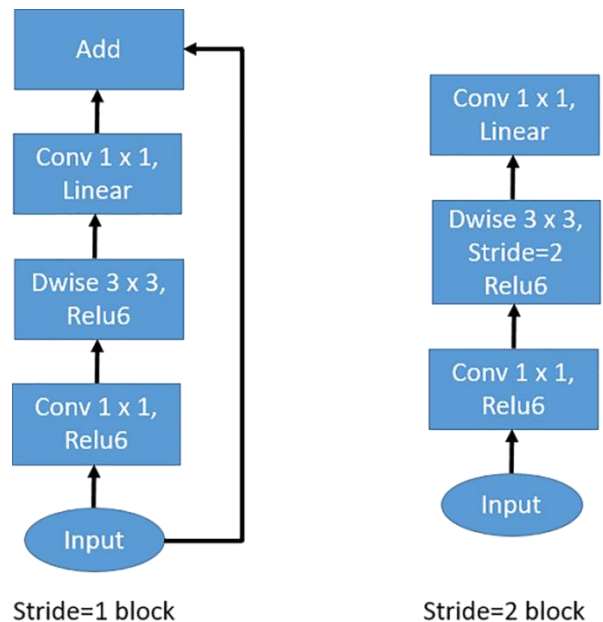

Figure2 illustrative example in graphical form of the MobileNetV2 architecture.

MobileNet v2 networks are well-suited for implementation on mobile terminals or embedded systems because to their small model size, cheap processing capacity, and high computation speed in comparison to more conventional convolutional neural networks. Moreover, it can keep up with the CPU's strict speed needs.

II. Transfer learning

It takes a significant amount of time and resources to train a deep neural network model on each job from the very beginning. The learning process is laborious and time consuming. In order for the network to begin its learning process, it must first texture the outline. Subsequently, the network must extract the characteristics after increasing the number of network layers. However, in the majority of instances, academics will use the information that they have gained in the past to apply to the current challenges that they face. Learning by transfer is the name given to this kind of information transmission (Zhuang et al., 2021). In the field of deep learning, transfer learning is combined with it, and the pre-training weight obtained by the model in similar tasks is used. This is done to avoid model training from scratch, reduce the cost consumption caused by the network model due to new learning, accelerate the network convergence speed, and further improve the stability and generalization ability of the model (Song et al., 2021).

The ImageNet data collection includes one thousand categories and encompasses one million two hundred thousand pictures. There is no shadow of a doubt that the utilization of pre-trained network models on such a massive data collection is capable of being successfully ported to a variety of picture categorization endeavors.

In this study, transfer learning is carried out by pre-training a deep CNN model using a portion of the ImageNet dataset. This portion of the dataset contains image classification data. A classifier was added to the pre-trained Mobilenet v2 model, and then it was used to download the private dataset and split it into 10 primary categories. The pre-trained model was imported via making use of the torch vision library, and samples from the private dataset were used for the training and testing of the model. By doing things this way, we were able to cut down on the time spent training the features extractor, which is what takes the most time for the majority of models, and instead train the classifier directly. Because of this, the amount of money spent on training has decreased.

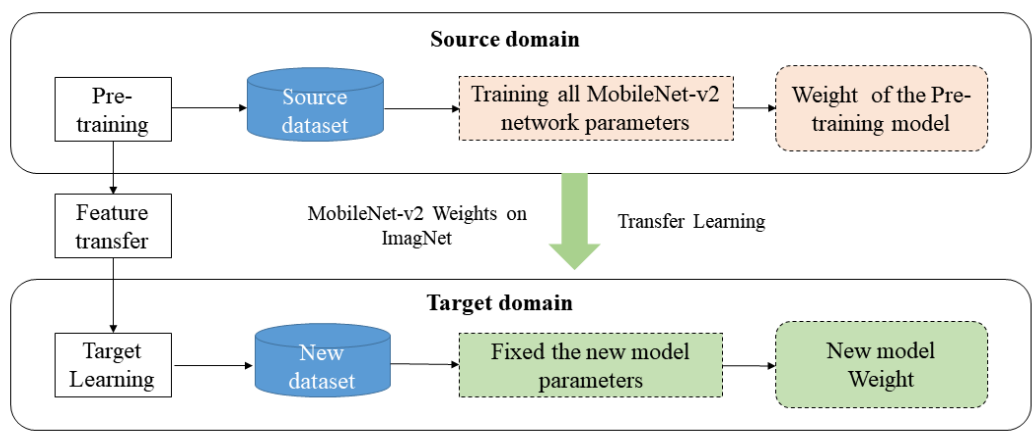

Figure3 transfer learning

## III. Convolutional block attention module

The Convolutional Block Attention Module, often known as CBAM, is a type of attention mechanism module that integrates Channel and Spatial. This module may be embedded into the Model module, trained end-to-end with Model, and only contributes a little amount of additional processing (Woo et al, 2018). Figure 4 demonstrates the Channel attention module in addition to the Spatial attention module. The Channel Attention Module is capable of taking an Intermediate Feature Map as an input and outputting both a 1-D Channel Attention Map and a 2-D Spatial Attention Map through the use of the CBAM. In order to aggregate the spatial information, the average pooling and maximum pooling methods are utilized. The approaches provide two descriptors, which are then passed to the same shared network in order to produce the channel attention map. The complete channel attention map can then be acquired by doing the following:

$$M_c(F) = \sigma\left(M(a(F)) + M(m(F))\right) = \sigma\left(W_1\left(W_0\left(F_{avg}\right)\right) + W_1\left(W_0\left(F_{max}\right)\right)\right) \text{ (formula 3)}$$

The spatial attention module can highlight the information region and generate two 2D maps, which are then linked and convolved by a standard convolutional layer to produce a 2D spatial attention map. This map is produced after first averaging and maximum pooling along the channel to generate efficient feature descriptors. In conclusion, the formulation for the output of the spatial attention mapping is as follows:

$$M_s(F) = \sigma\left(f\left(f_c\left(F_{avg}, F_{max}\right)\right)\right) \text{ (formula 4)}$$

F is the input feature graph, is the sig-moid nonlinear activation function, M is the forward calculation function of multi-layer perceptron without bias, a and m are the mean and maximum pooling functions respectively, W0 and W1 are the weights of 2 linear layer, and Favg and Fmax are the mean and maximum pooling functions respectively. In the equation that was just presented, F represents the input feature graph, represents the sig-moid nonlinear.

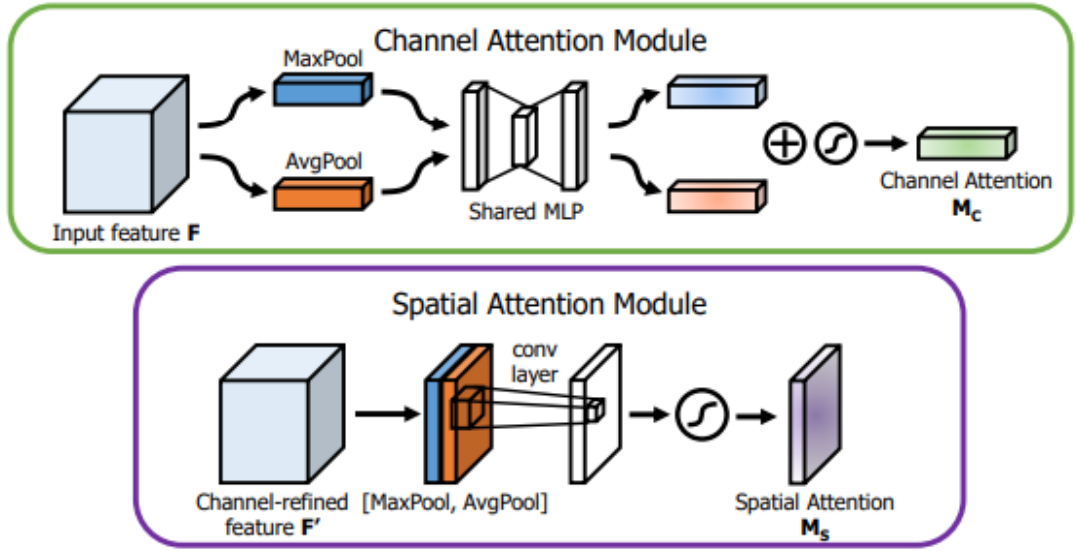

Figure4 Channel attention module and Spatial attention module

It is necessary to arrange the Channel attention module and the Spatial attention module in a sequential fashion in order to generate the CBAM. This provides the model with the ability to concentrate on important features in both the channel and spatial dimensions while suppressing features that are not necessary. Figure 5 provides a high-level view of CBAM, while the following provides a concise summary of the complete calculating process:

$$F' = M_c(F) \otimes F \text{ (formula5)}$$
$$F'' = M_s(F') \otimes F \text{ (formula6)}$$

It is necessary to arrange the Channel attention module and the Spatial attention module in a sequential fashion in order to generate the CBAM. This provides the model with the ability to concentrate on important features in both the channel and spatial dimensions while suppressing features that are not necessary. Figure 5 provides a high-level view of CBAM, while the following provides a concise summary of the complete calculating process:

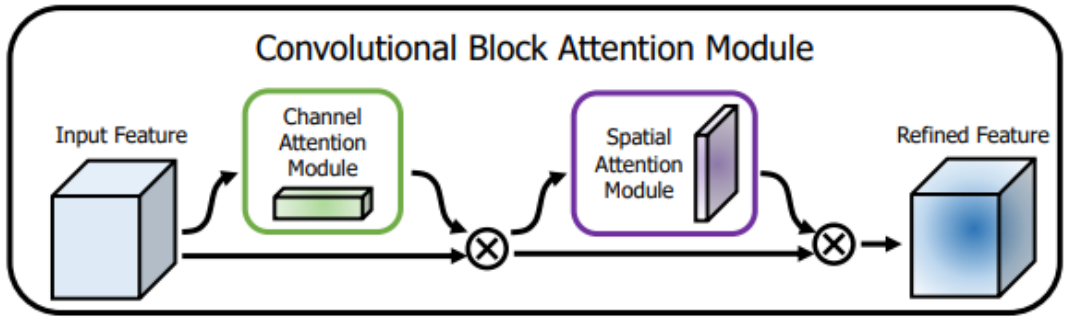

Figure5 The overview of CBAM

This algorithm integrates the attention mechanism in the dimension of the channel and the space, strengthens the important features while suppressing the unimportant features, and further highlights the required features in order to improve the classification accuracy of the model. The CBAM structure is introduced in the first layer of the Mobilenet v2 network by this algorithm. Because CBAM is a lightweight module, it does not have a significant impact on the total number of model parameters, but it does improve emphasis on the model's primary features. This effectively guarantees that the benefits of a lightweight model will be realized.

IV. Perceptual hash algorithm

The perceptual hash algorithm (Samanta & Jain, 2021) begins by reducing the size of the picture and simplifying the colors. Next, it aggregates the decomposition frequency and trapezoidal shape of the picture by using the discrete cosine transformation method. After that, it reduces the DCT, keeps the 8*8 matrix in the upper right corner, calculates the average value of all 64 values, further reduces the DCT, and finally calculates the hash value. The algorithm creates a unique string that acts as a "fingerprint" for each image, and it then analyzes the results by contrasting the various "fingerprint" strings. When the result is closer, it indicates that the picture is more comparable. The advantage of using this approach is that regardless of whether the height, width, brightness, or color of the image is modified, the result value of the hash value will not remain intact. This allows one to prevent the impact that would be caused by the modest modification of the image. An online image identification system is constructed in this research using the perceptual hash method and the concept of searching for images using photos.

V. Measurement

There is currently no standardized dedicated data collection that can be used for the categorization of photos of lost and found items. The study of confidential data sets is the focus of this work. Using the results of the poll, streamline the complicated lost and found categories into the top 10 categories that are the easiest to lose track of, with a total of 10499 photographs. In addition, in order to match the conditions of a practical application, we have not performed any pre-processing on the dataset. Directly scaling the collected lost and found data set pictures to the input size of the network model will directly lead to the loss of some information in the picture, or even direct distortion, and will ultimately directly affect the image classification effect and recognition accuracy. This is because the original image had an uneven quality and a high resolution. As a result, prior to the pre-training, this article automates the data set photos in batches by using the design action in Adobe Photoshop software. Additionally, the article standardizes the image resolution to 300 x 300, which helps to minimize the overall picture size. The data processing of the sample data has to be improved in this work in order to increase the capacity of the model to generalize its findings. The 10,499 common lost and lost photographs augmented by the data were divided into the training set and validation set with 70% and validation set with 30%.

**Table 1. Sample distribution of the lost object Figure slice dataset used in this paper**

| Class name | Number of images |
|:---:|:---:|
| bag | 1036 |
| book | 1067 |
| card | 1043 |
| earphone | 1025 |
| key | 1070 |
| lipstick | 1066 |
| Phone | 1068 |
| umbrella | 1045 |
| USBflashdisk | 1020 |
| vacuumcup | 1059 |

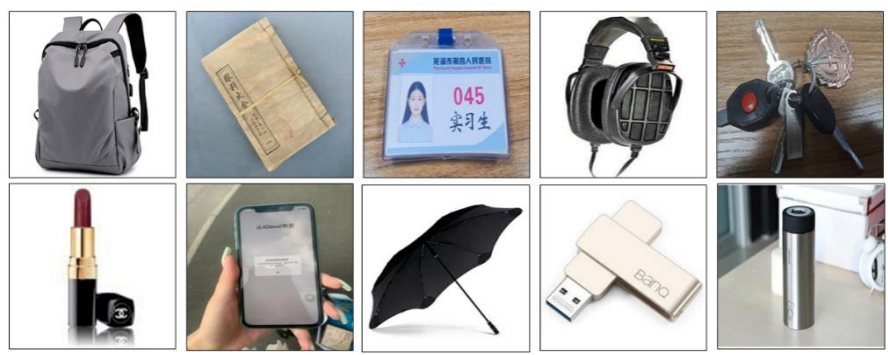

Figure6 The data inside the dataset

This article employs Accuracy, AgePrecision (AP), Recall, Precision, and Loss as model performance evaluation indicators in order to analyse the advantages and disadvantages of the model.

Calculating values using typical complex matrix approaches like as accuracy, recall, and precision may be difficult. When it comes to image classification, the confusion matrix is most commonly employed to make a comparison between the classification and the actual measurement value. This allows for a more understandable and accurate description of the correctness of model categorization. The formula for the computation is as follows:

$$\text{Accuracy} = \frac{TN+TP}{TN+FN+FP+TP}$$ （formula7）

$$\text{Recall} = \frac{TP}{TP+FN}$$ （formula8）

$$\text{Precision} = \frac{TP}{TP+FP}$$ （formula9）

The number of successfully predicted positive samples is denoted by the letter TP in the formula 7,8,9, the number of correctly anticipated negative samples is denoted by the letter TN. FN is for the number of samples that had errors predicted for them when they were positive, and FP stands for the number of samples that had errors predicted for them when they were negative.

The performance of the model recognition accuracy of the selected model may be evaluated in an objective manner by using the average accuracy rate. The mathematical formula for the calculation is as follows:

$$AP = \frac{\sum_1^E \text{Accuracy}}{E}$$ （formula10）

The term AP, which is utilized in the aforementioned formula 10, stands for the average accuracy of the model that was picked, while E stands for the total number of iterations, and Accuracy stands for the accuracy of each iteration of the model that was selected.

The degree to which the model's predicted value and the actual value deviate from one another may be estimated using the loss value. If the loss value is lower, then the model's predicted result will be closer to the actual outcome. The formulae for the calculations are as follows:

$$LOSS = \frac{1}{N}\sum_i L_i = \frac{1}{N}\sum_i -\sum_{c=1}^K y_{ic} \log(p_{ic})$$ （formula11）

In the previous formula number 11, Loss is a numerical representation of the loss value associated with the model that was chosen, and K is the total number of categories. pic is the anticipated probability that sample I belongs to category c. If the category is the same as the category of sample I it has a value of 1, and otherwise, it has a value of 0. The cross-entropy loss function is a convex function, and it is possible to calculate the value that is best for the whole world.

**Experimental results:**

I. Training

In the course of our research, we looked at the problem of lost-object picture categorization. In order to classify images, we made use of Convolutional block attention module in conjunction with the well-known framework MobileNet v2. We utilized this architecture (transfer learning) to speed up the learning process while also reducing the amount of time needed for training. In order to evaluate the effectiveness of transfer deep learning, we examined the outcomes of three different optimizer functions: the Adaptive Moment Estimation Algorithm (ADAM), the Root Mean Square Propagation (RMSprop), and the Stochastic gradient descent (SGD) algorithm. We noticed that the optimizer SGD function functioned in a better method and got an accuracy of 96.8 percent, while the loss detected was 0.0047. We figured this out by looking at the table. The results of the SGD function's performance are presented in the form of a linear graph of the accuracy and loss parameters in the figure. When compared to the other optimizers, SGD's training time for the model was significantly shorter, despite the fact that all other model parameters remained same.

**Table2. Comparative chart of results using optimizer function**

| Components | SGD | RMSprop | ADAM |
|---|---|---|---|
| Accuracy/% | 96.8 | 89.5 | 94.7 |
| Loss | 0.0047 | 0.0050 | 0.0048 |
| Freeze Epoch | 50 | 50 | 50 |
| Freeze batch size | 32 | 32 | 32 |
| UnFreeze Epoch | 400 | 400 | 400 |
| Unfreeze batch size | 64 | 64 | 64 |
| Init lr | 1e-2 | 1e-2 | 1e-2 |
| Lr decay type | cos | cos | cos |
| momentum | 0.9 | 0.9 | 0.9 |
| Execution Time/ minutes | 5.38 | 14.28 | 10.71 |

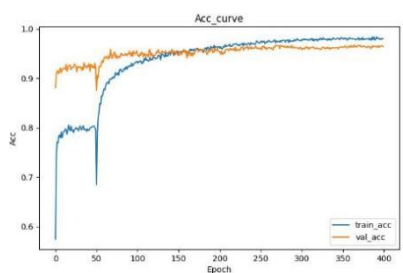
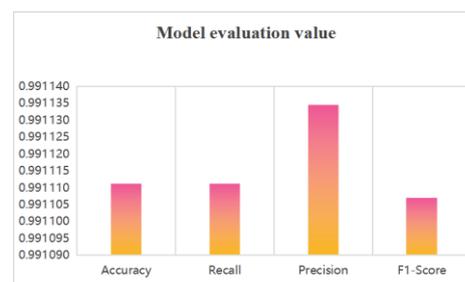

Figure7a(left) Training and validation curve of the model

Figure7b(right) the valuation index of different measure index

It is abundantly evident, after reviewing the preceding explanation complete with graphs and table, that the optimizer function SGD performed far better than the performance of the other functions when applied to the issue of lost-object picture classification.

EfficientNet, Inception v4, ViT-B. 32, DenseNet201, VGGNet19, ShuffleNet v2, ResNet152 and MobileNet v3 were the eight transfer learning algorithm models of the same kind that were chosen at random for the purpose of doing a comparison study in order to test the efficiency and superiority of the research approach that was provided in this work.The optimal parameter values of the model convergence were selected as the respective parameters.This study's private data set is used to train and verify a total of nine models using these settings. The outcomes of the training are presented in Table 3.

**Table3. Results of the selected model training**

| Model | Accuracy/% | AP/% | Total parameters/M |
|---|---|---|---|
| EfficientNet（Koonce et al，2021） | 93.9 | 89.0 | 66.348 |
| Inception v4（Szegedy et al，2017） | 89.5 | 80.3 | 41.158 |
| ViT-B/32（Dosovitskiy et al，2010） | 94.9 | 94.3 | 104.766 |
| DenseNet201（Huang et al，2017） | 95.9 | 95.4 | 20.014 |
| VGGNet19（SIMONYAN et al，2021） | 95.6 | 94.7 | 139.611 |
| ShuffleNet v2（Ma et al，2018） | 88.6 | 82.3 | 2.279 |
| ResNet152（He et al，2016） | 87.8 | 81.9 | 60.193 |
| MobileNet v3（Koonce et al，2021） | 95.7 | 94.9 | 5.483 |
| us | 96.8 | 96.2 | 3.505 |

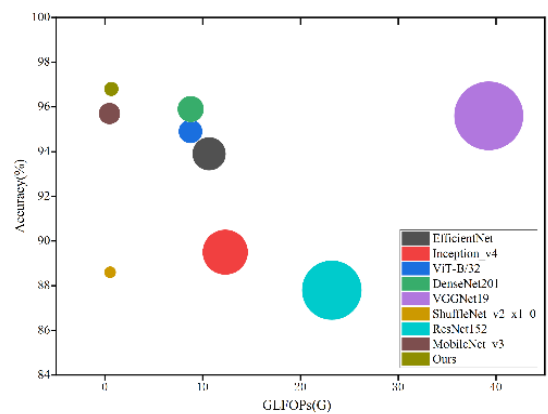

Figure8 The performance, accuracy and parameter amount compare of difference models

The algorithm that is suggested in this article is transfer learning, which is based on the improved MobileNetv2. This can be seen in Table 3. The results of the experiments demonstrate that the method under study has an average accuracy of 96.2% when applied to data sets that were self-built. This is a higher accuracy rate than EfficientNet, Inception v4, ViT-B. 32, DenseNet201, VGGNet19, ShuffleNet v2, ResNet152 and MobileNet v3 correspondingly. 7.2%, 15.9%, 1.9%,0.8%,1.5%, 13.9%, 14.3% and 1.3% of the total learning was transferred from other models. The suggested technique has obtained the greatest accuracy rate in private data sets, which is 96.8%, along with strong generalization ability and resilience. When compared to the similar kind of transfer learning model that has been developed, this is how it stands out. In correlation with the information shown in Figure 8, The ordinate in the figure shows the test accuracy, the abscissa represents GFLOPs, the circle colors reflect distinct transfer learning models, and the figure size denotes total parameters. The suggested method gets the maximum accuracy on private datasets, and it is robust and has a decent ability to generalize. When Figure 8 and Table 3 are considered together, a further complete evaluation of the performance of each model is carried out. This model is only second to MobileNet v3 and ShuffleNet v2, but it is worth recognizing that the Total parameters of the proposed model are almost half that of MobileNet v3 Total parameters, and the accuracy is higher than ShuffleNet v2 8.2%. MobileNet v3 and ShuffleNet v2 are the only models that are ahead of this model. In comprehensive comparison, the model that is being offered is a fantastic lightweight network, which provides the possibility that the model might be transplanted to mobile devices.

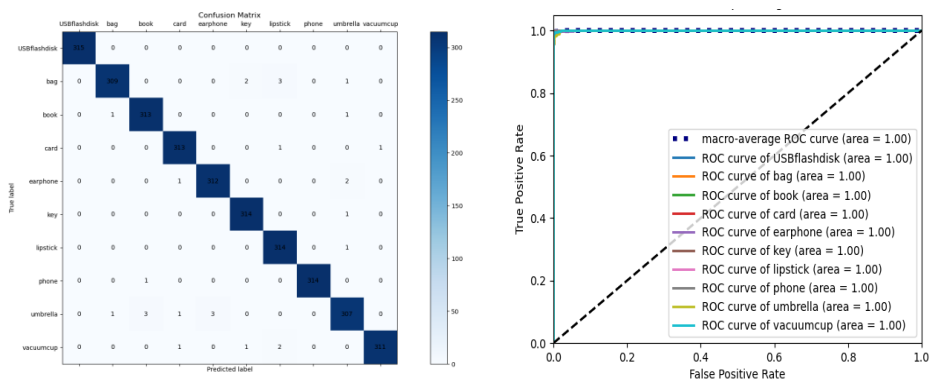

Figure9a(left) Confusion matrix of the model

Figure9b(right) ROC curve of the model

We decided to combine the confusion matrix from the model single-picture test with the ROC curve for the study so that we could have a more in-depth look at the data. The confusion matrix is the method for measuring the accuracy of classification models that is the simplest and most straightforward. Calculating the confusion matrix of the suggested model requires using equation (7) and equation (9). This matrix represents the identification result of the different types of lost object photographs and is derived from these equations. The confusion matrix is shown in Figure 9a, with the rows representing the expected category for the item that was lost, and the columns representing the actual category for the object that was lost. Figure 9b illustrates the characteristic curves of the class 10 lost item categories for the enhanced MobileNetv2 transfer learning model. ROC stands for receiver operating characteristic. The fact that the curves for all ten classes are located in the top left corner is evidence that the model is accurate, and the fact that the area under the ROC curve for each class is equal to one further demonstrates the superiority and efficiency of the model that was provided.

II. Inference

In order to successfully apply the model to the actual world, it needs to be simple enough that it can be executed on most laptops. In order to accomplish this, we implement this model as well as a model that is very close to it on a regular computer that has a CPU, and we comprehensively consider the inference speed of the computer as the evaluation index. It takes the model 1.5 seconds to reason per picture on the CPU, which is significantly longer than the standard migration model of this type or even 1/16 of the EfficientNet's processing time. This can be seen in Figure 10a. Because it is such an exceptionally lightweight network, it is not hard to imagine that the concept will eventually be adapted to work with other mobile devices in the not too distant future. In light of this, we provide an engineering system that integrates the perceptual hashing method.

This paper, which is based on the model training using the pytorch and torchvision machine learning frameworks, uses the Spring Boot framework as the back-end service of the web page in order to realize the online recognition of pictures. The front-end web page uses the Layui framework in order to port the trained model to the Web end in order to realize the online recognition of images that are uploaded by users. The steps involved in the online identification procedure are as follows: the user launches the browser, navigates to the system's website, navigates to the system's home page, clicks the button to upload the local image, the front end sends the POST request to the back end, and the image is transmitted to the back end. After the back end has received the picture that was uploaded by the user, it will first make a request to the trained and upgraded MobileNet v2 migration learning model for prediction and recognition, and it will then return a category. After it has determined the category, it will next submit a request to the database, asking it to deliver an array containing the picture address associated with the category it has just determined. After obtaining the address, the image is then downloaded by using the image address, and then the perceptual hash algorithm is used to compare the downloaded image to the image that was uploaded by the user. The alignment will return a number for the similarity error, and the array will be used to send the few photos that have the least significant value for the similarity error to the front end. Figure 10b depicts the engineering interface after the model has been applied to the situation.

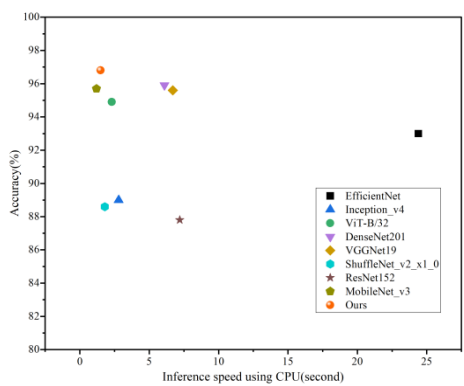
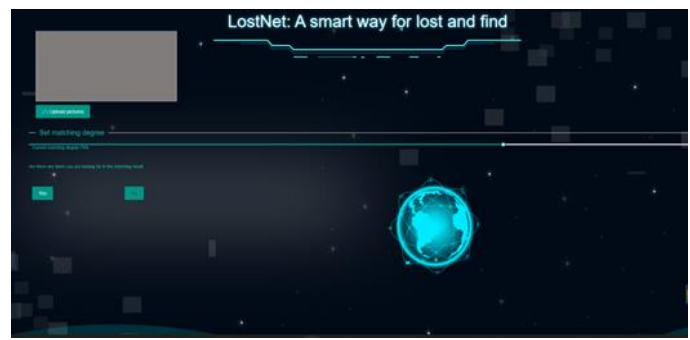

Figure10a(left) The speed of inference of each model in cpu

Figure10b(right) The improved model is applied to the actual engineering interface diagram

**Concludes:**

This research presents a novel design plan in light of the difficulties that are associated with the management of urban traffic operators. Even while the method of merging convolutional neural networks with transfer learning has caused quite a stir in the world of image recognition, it is not widely utilized when it comes to the automatic recognition of photos that have been misplaced or found. This study offers a new technique of intelligent picture identification based on hash algorithm to give full play to the advantages of hash algorithm. The approach makes use of the concept of "search by map," and it is based on hash algorithm. Because there are a significant number of missing images in the library and a great number of feature points that have been extracted, the CBAM structure has been introduced in the first layer of the network. This has been done in order to integrate the attention mechanism in the dimension of channel and space in order to further highlight the required features, which will ultimately lead to an improvement in the classification accuracy of the model. Improved MobileNetv2 is used to establish a transfer learning training model to identify common lost objects. This is done in order to reduce the ncum of ways in which lost objects and database pictures can be compared to one another, thereby making it possible to search for objects in a more convenient manner. Extensive experimental arguments on photos taken from the loss-object dataset, employing a variety of transfer learning models. The findings indicate that the proposed model has a high recognition accuracy in the GFLOPs and Total parameters extremes, and the features derived by this model significantly beat those extracted by the other approaches when it comes to the classification job. In order to enhance the precision with which missing object categories may be identified, the proposed model is being developed and put into practice in the search sector.

However, further solutions are required in order to further increase the accuracy of the model and the segmentation of the various kinds of lost object. In order to do this, the following stage will involve further subdividing each category so that it can be utilized in a wider variety of search circumstances.

**Acknowledgement:**

This work was supported by Provincial College Students' Innovation and Entrepreneurship Project Project for College Students [Grant numbers S202110368112]; University Humanities and Social Science Research Program of Anhui Province [Grant numbers SK2020A0380]; School level Project of Key Humanities and Social Sciences Research Base of Anhui Province, Center for Mental Health Education of College Students [Grant numbers SJD202001]; and School level Project of the Young and Middle-Aged Natural Science Foundation of Wannan Medical College[Grant numbers WK202115]